\pdfoutput=1
\documentclass[lettersize,journal]{IEEEtran}
\usepackage{amsmath,amsfonts}
\usepackage{algorithmic}
\usepackage{array}

\usepackage{textcomp}
\usepackage{stfloats}
\usepackage{url}
\usepackage{verbatim}
\usepackage{graphicx}

\usepackage{color}
\usepackage{makecell}
\usepackage{multirow}
\usepackage{multicol}
\usepackage{enumitem}
\usepackage{booktabs}

\usepackage{hyperref}
\usepackage{caption}
\usepackage{subfigure}
\usepackage[normalem]{ulem}

\def\BibTeX{{\rm B\kern-.05em{\sc i\kern-.025em b}\kern-.08em
    T\kern-.1667em\lower.7ex\hbox{E}\kern-.125emX}}
\usepackage{balance}
\begin{document}

\title{Enhancing Textual Personality Detection toward Social Media: Integrating Long-term and Short-term Perspectives}

\author{
	\IEEEauthorblockN{Haohao Zhu$^{a}$, Xiaokun Zhang$^{a}$, Junyu Lu$^{a}$, Youlin Wu$^{a}$, Zewen Bai$^{a}$, Changrong Min$^{b}$, Liang Yang$^{a}$, Bo Xu$^{a}$, Dongyu Zhang$^{c}$, Hongfei Lin$^{a*}$} \\
    \IEEEauthorblockA{$^a$ School of Computer Science and Technology, Dalian University of Technology, Dalian, China} \\
    \IEEEauthorblockA{$^b$ Criminal Investigation Police University of China, Shenyang, China} \\
    \IEEEauthorblockA{$^c$ School of Foreign Languages, Dalian University of Technology, Dalian, China} \\
    \IEEEauthorblockA{\{zhuhh, dutljy, wuyoulin, dlutbzw\}@mail.dlut.edu.com, \{liang, xubo, zhangdongyu, hflin\}@dlut.edu.cn}, \{dawnkun1993, mcr19940816\}@gmail.com
}

\maketitle

\begin{abstract}
Textual personality detection aims to identify personality characteristics by analyzing user-generated content on social media platforms. 
Extensive psychological literature highlights that personality encompasses both long-term stable traits and short-term dynamic states. 
However, existing studies often concentrate only on either long-term or short-term personality representations, neglecting the integration of both aspects.
This limitation hinders a comprehensive understanding of individuals' personalities, as both stable traits and dynamic states are vital. 
To bridge this gap, we propose a \textbf{D}ual \textbf{E}nhanced \textbf{N}etwork (DEN) to jointly model users' long-term and short-term personality traits. 
In DEN, the Long-term Personality Encoding module models stable long-term personality traits by analyzing consistent patterns in the usage of psychological entities.
The Short-term Personality Encoding module captures dynamic short-term personality states by modeling the contextual information of individual posts in real-time.
The Bi-directional Interaction module integrates both aspects of personality, creating a cohesive and comprehensive representation of the user’s personality.
Experimental results on two personality detection datasets demonstrate the effectiveness of the DEN model and underscore the importance of considering both stable and dynamic aspects of personality in textual personality detection.
\end{abstract}


\section{Introduction}

Textual personality detection toward social media aims to identify an individual's personality characteristics reflected in their written text \cite{yang2023orders}.

By analyzing a user's personality characteristics, we can gain a deeper understanding of their preferences \cite{furnham1981personality}, behavior patterns \cite{schweiker2016influence}, and decision-making processes \cite{gambetti2019personality}. As a result, personality detection has been widely applied in various downstream tasks, such as personalized advertising and recommendation systems \cite{dhelim2022survey}, dialogue systems \cite{yang2021improving}, and individual career planning \cite{kern2019social}.

\begin{figure}[!htb]
\centering
    \includegraphics[scale=0.6, trim=75 140 0 620, clip]{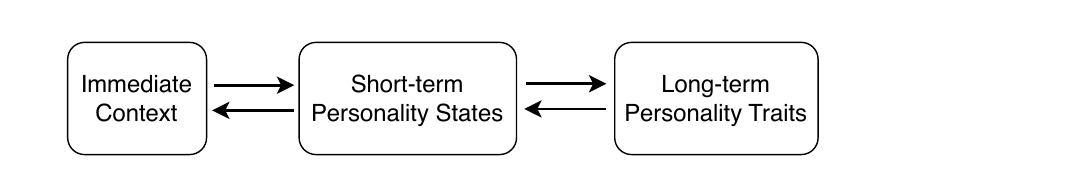}

\caption{Personality encompasses long-term, relatively stable internal traits while also undergoing dynamic changes and development influenced by various short-term states within the immediate context.}
\label{fig:personality development}
\end{figure}

Traditionally, personality assessment relies on questionnaires designed by psychologists, which can be time-consuming and labor-intensive, and impractical for the online world. 
The rise of social media has created vast opportunities for automated textual personality detection, attracting a growing number of researchers in the field \cite{fang-etal-2023-text}.

Textual personality detection, although resembling a multi-label classification task, poses unique challenges. For example, the input comprises multiple posts rather than a single text \cite{yang2023orders}. Each post can reflect the user's current short-term personality state, while the collection of all posts can capture the user's long-term trait. To address these challenges, numerous studies have been conducted, broadly categorized into two main types. 

On the one hand, some methods utilize a flat structure and focus on modeling the overall long-term representation of users to capture their stable personality traits for personality detection. This can be accomplished by utilizing various linguistic or statistical features \cite{pennebaker2001linguistic,zhang2010understanding} or concatenating the user's post representations into a single representation \cite{jiang2020automatic} to capture the long-term representation. 

On the other hand, another line of research, employing a hierarchical structure, involves first modeling the current short-term representation of users' states using their individual posts. Subsequently, these representations are aggregated to obtain an overall user representation for personality detection \cite{lynn2020hierarchical,yang2021multi,yang2023orders}.

However, the aforementioned two types of approaches may be insufficient in effectively combining the user's current short-term states representation with their relatively stable long-term traits representation. Indeed, numerous psychological studies have highlighted that a user's personality encompasses their long-term relatively stable internal traits, while also undergoing dynamic changes and development influenced by various short-term states that within the immediate context \cite{collins1990social, wagner2016personality, denissen2019transactions, anaya2019personality, lee2019study, fang-etal-2023-text}, as illustrated in Figure \ref{fig:personality development}.

\begin{figure}[ht]
\centering
    \includegraphics[scale=0.25, trim=0 0 0 0, clip]{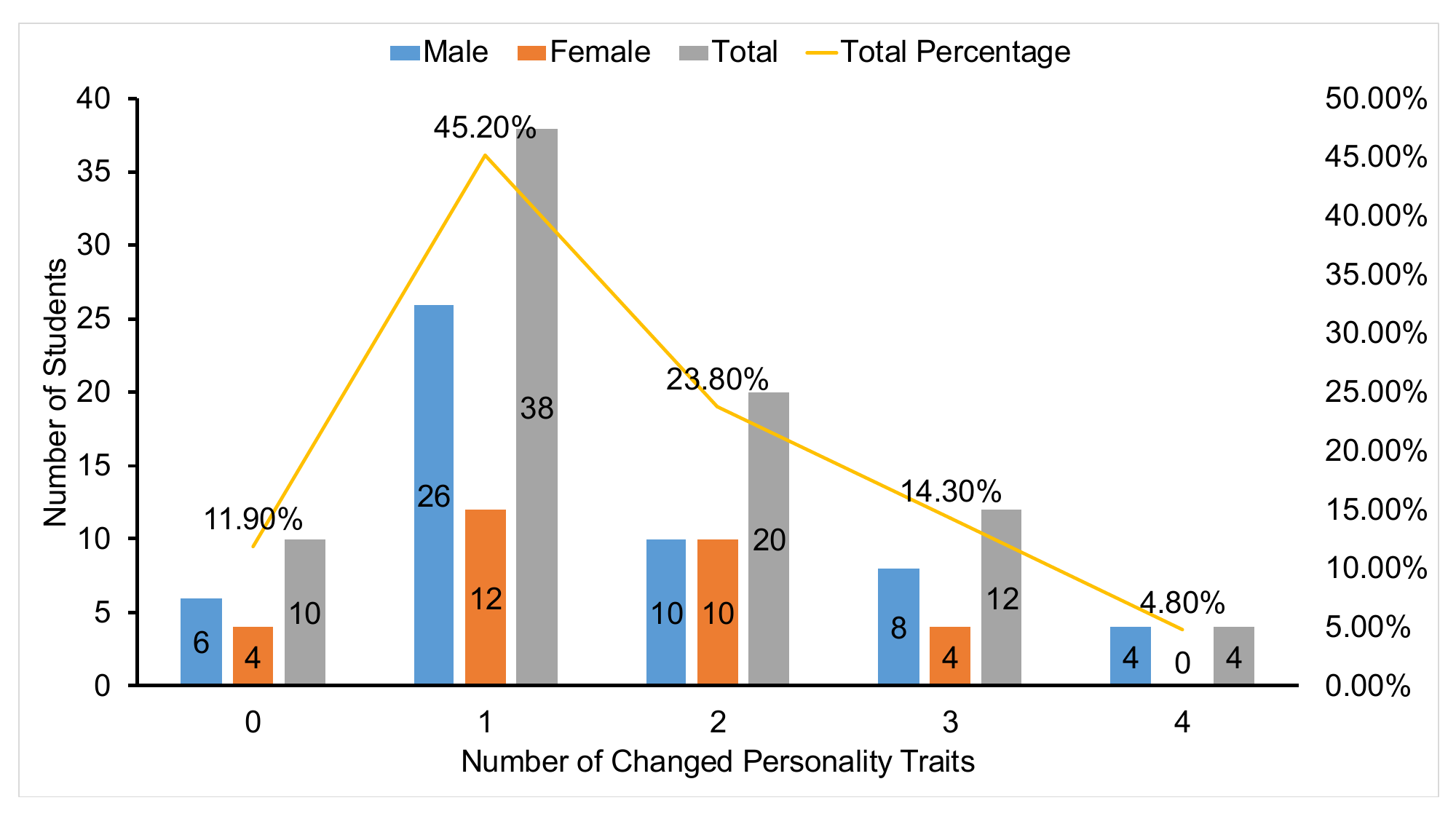}

\caption{Distribution of changes in MBTI personality characteristics among 84 college students over a 6-year period (from the first year of pre-university to the fourth year of undergraduate study). The bar chart compares the number of personality changes among different gender groups, while the line graph illustrates the trend of the percentage of different personality change counts. Data are from \cite{lee2019study}}
\label{fig:Personality Changes}
\end{figure}

Specifically, \cite{lee2019study} conducted a study on 84 college students, who are the primary users of social media, to investigate the change of personality types focusing on the Myers-Briggs Type Indicator(MBTI) \cite{myers1987introduction}. The MBTI categorizes personality types into four traits: extraversion/introversion, sensing/intuition, thinking/feeling, and judging/perceiving. From their results depicted in Figure \ref{fig:Personality Changes}, we can find that on the one hand, only 4.8\% of students experienced a complete change in their personality indicating that personality is relatively stable. Additionally, more than half of the students who underwent personality changes only changed one personality trait, further supporting the notion of stability in personality.
On the other hand, we also found that a large majority of the students (88.1\%) experienced changes in at least one indicator of personality. This highlights the dynamic nature of personality and suggests that while certain traits may remain relatively stable, other aspects of an individual's personality can undergo significant changes over time.

The above findings emphasize the complexity of personality development, with a combination of stability and change occurring simultaneously. It underscores the importance of considering both the stable long-term traits and the dynamic short-term states when studying and predicting personality.

Unfortunately, it is non-trivial to capture both the long-term and short-term perspectives of user personality due to the following challenges: (1) Effectively modeling long-term personality representations, which reflect stable and consistent traits over time; (2) Accurately modeling short-term personality states, which are dynamic and context-dependent; and (3) Integrating these two distinct representations to enhance personality prediction.

To address these challenges, we propose the \textbf{D}ual \textbf{E}nhanced \textbf{N}etwork (DEN) for personality detection, which jointly captures long-term and short-term personality representations while dynamically integrating them to enhance personality detection. 
Specifically, DEN introduces three novel modules:
(1) \textit{Long-term Personality Encoding}: This module models users’ long-term stable personality traits by aggregating their consistent usage patterns of psychological cues over time.
It achieves this by constructing a user-specific psychological graph using LIWC (Linguistic Inquiry and Word Count) \cite{pennebaker2001linguistic}.
By applying graph neural network (GNN) to this graph, this module effectively analyzes the relationships between psychological entities, learning insightful representations of enduring personality characteristics.
(2) \textit{Short-term Personality Encoding}: This module models users’ short-term dynamic personality states by capturing their immediate linguistic expressions in individual posts. 
Leveraging a pre-trained language model with multi-head self-attention mechanism, it encodes each post to reflect the user’s transient personality changes, enabling effective representation of short-term personality states.
(3) \textit{Bi-directional Interaction}: This module uses a bipartite heterogeneous graph to facilitate interaction between long-term and short-term representations. 
Through this interaction, both perspectives are mutually refined, resulting in a more comprehensive and accurate personality representation.

Extensive experiments demonstrate the effectiveness of DEN in the personality detection task. 
The key contributions of this paper can be summarized as follows:

\begin{enumerate}
    \item This research is the first to delve into the impact of both the dynamic and stable nature of personality characteristics on the process of personality detection, providing a novel perspective on this task.
    \item We introduce DEN, a novel approach that can simultaneously capture both the long-term and short-term personality representations, and effectively integrate them for personality detection.
    \item We conduct extensive experiments on two benchmark datasets, demonstrating the promising performance of our DEN and providing insights into the effects of each key module.
\end{enumerate}

\section{Related Work}

The advent of the internet and social media platforms has led to a wealth of user-generated text data, prompting researchers to explore methods for extracting user information from these textual sources \cite{fang-etal-2023-text}. Textual personality detection has emerged as an important problem in this context, and it has garnered growing interest among researchers \cite{xue2018deep,lynn2020hierarchical,yang2023orders}.

According to the distinction between long-term and short-term personality characteristics, research on textual personality detection can be categorized into two main types. 

One type of research in textual personality detection utilizes a flat structure, focusing on modeling the overall long-term representation of users, which reflects their stable personality traits. Various approaches have been proposed in this category. For instance, studies such as \cite{mehta2020bottom, lin2023novel, kumar2023personality} employ various linguistic features extracted from the entirety of users' posts to capture their language habits and represent their long-term traits. Similarly, \cite{zhang2010understanding, cui2017survey} utilize statistical features to model users' overall language patterns and represent their long-term traits. Some approaches, like \cite{jiang2020automatic} concatenate users' entire texts, while others, like \cite{yang2021multi, jain2022personality} concatenate representations from different texts to learn the overall user representation. 
These methods primarily focus on capturing the consistent personality traits that persist over an extended period. However, they may not fully capture the dynamic changes and fluctuations in a user's personality that can occur in short-term states.

On the other hand, the second category of methods adopts a hierarchical strategy, focusing on learning the current short-term representation of users' short-term states using their current posts. These short-term representations are then aggregated to form an overall user representation for personality detection.
For example, \cite{lynn2020hierarchical} employed a hierarchical attention network \cite{yang2016hierarchical} and \cite{majumder2017deep} employed hierarchical convolutional neural network (CNN) to generate personality representations from posts for personality detection. 
\cite{yang2021psycholinguistic, yang2023orders, zhu2024data} utilized graph neural networks to aggregate text representations. 
These approaches capture short-term variations in personality by modeling dependencies within each post and aggregating representations across multiple posts to create an overall user representation. However, while effective in capturing short-term variations and dependencies within each post, they may overlook the enduring traits that remain consistent over an extended period, thus not fully capturing the long-term stable aspects of personality.

Indeed, the aforementioned methods have significantly improved the effectiveness of automatic personality detection from various perspectives.
However, they all encounter challenges in simultaneously modeling both the long-term and short-term aspects of personality and effectively integrating them for personality detection.

\section{Methodology}

\subsection{Problem Definition}

Textual personality detection task can be defined as a multi-document, multi-label classification task \cite{lynn2020hierarchical, yang2023orders}. 
Formally, we are given a set of user-generated posts denoted as $P = \{p_1, p_2, \ldots, p_N\}$, where each $p_i = \{t_{1i}, t_{2i}, \ldots, t_{Li}\}$ represents the $i$-th post consisting of $L$ tokens. 
The objective is to predict personality characteristics represented as $Y = \{y_1, y_2, \ldots, y_T\}$, where each $y_i \in \{0, 1\}$, and $T$ is the number of personality characteristics. 
For example, in the MBTI taxonomy, there are four personality characteristics represented by $T = 4$, while in the Big-Five taxonomy, there are five personality characteristics represented by $T = 5$. In this paper, we use the MBTI datasets for validation, hence $T = 4$.
In addition, to learn users' long-term personality representations, we incorporate the external LIWC Psychological Knowledge Base \cite{pennebaker2001linguistic}, which contains various psychological entities and their corresponding categories, as part of the input.

\subsection{Overview of DEN}

\begin{figure*}[!ht]
\centering
    \includegraphics[scale=0.85, trim=0 205 0 110, clip]{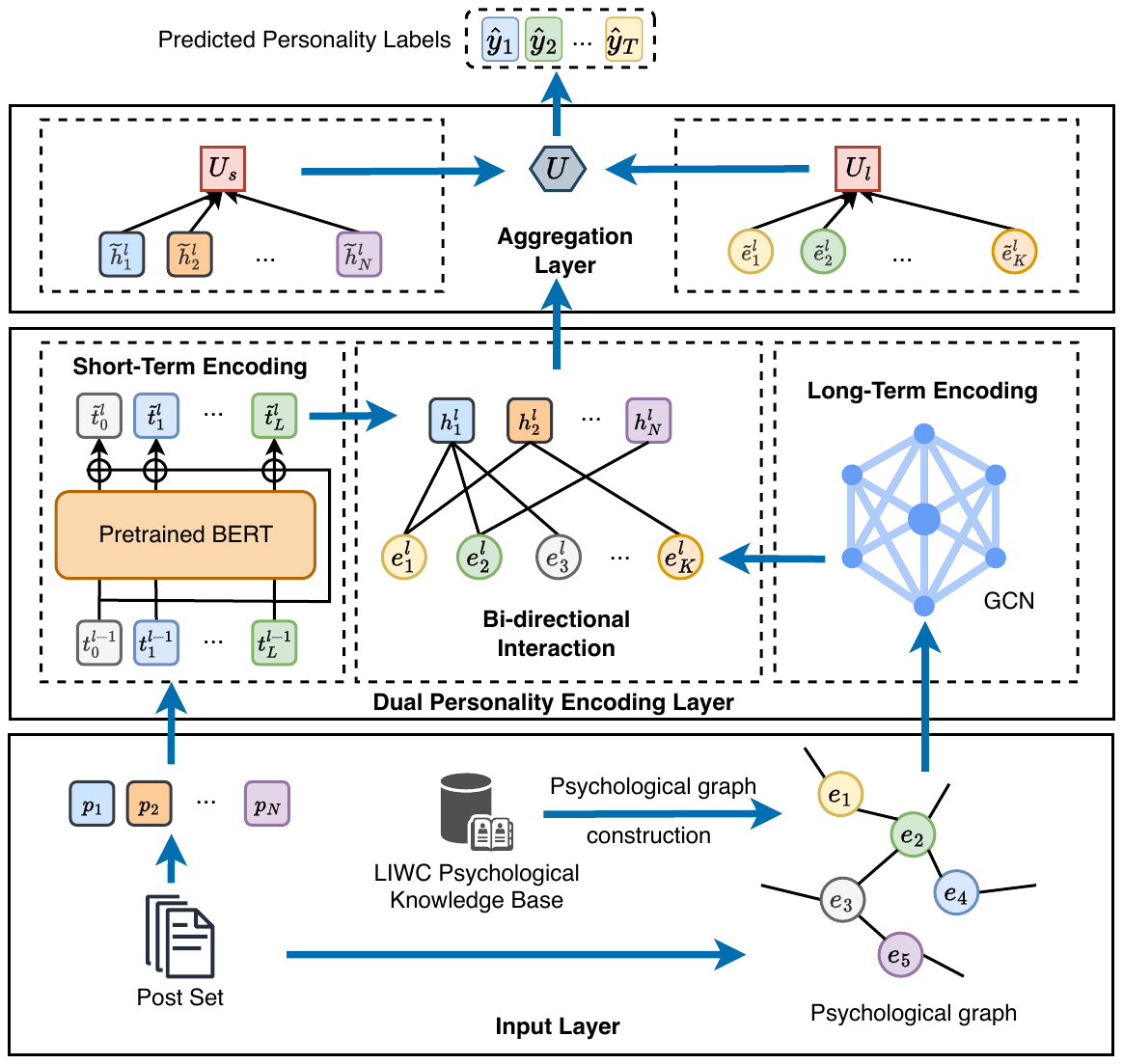}
\caption{The architecture of the proposed DEN model, which consists of three layers: Input Layer, Dual Personality Encoding Layer and Aggregation Layer. The Dual Personality Encoding Layer comprises three key modules: Long-term Personality Encoding, Short-term Personality Encoding, and Bi-directional Interaction.}
\label{fig:model}
\end{figure*}

The overall architecture of the proposed DEN model is depicted in Figure \ref{fig:model}. It consists of three layers: Input Layer, Dual Personality Encoding Layer and Aggregation Layer. Specifically, the Dual Personality Encoding Layer comprises three key modules: Long-term Personality Encoding, Short-term Personality Encoding, and Bi-directional Interaction. The Input Layer is in charge of preprocessing the data to create input for the subsequent Dual Personality Encoding Layer. The Dual Personality Encoding Layer is designed to model long-term stable personality traits, capture users' short-term personality states, and facilitate dynamic interactions between these two aspects. Finally, the Aggregation Layer aggregates the representations of both long-term and short-term traits to generate the ultimate user representation for the final personality characteristics classification. Next, we will provide a detailed explanation of each component of DEN.

\subsection{Input Layer}

The Input Layer is responsible for acquiring the inputs for the Dual Personality Encoding Layer from user-generated posts $P = \{p_1, p_2, ..., p_N\}$ and an external LIWC Psychological Knowledge Base \cite{pennebaker2001linguistic}. Specifically, we need to prepare input for the Long-Term User Encoding module, which requires a psychological graph, denoted as $\mathcal{G}(M, A)$. This graph comprises a entity node feature matrix $M \in \mathbb{R}^{K \times d_{g}}$ representing entity embeddings, and an adjacency matrix $A \in \mathbb{R}^{K \times K}$ that captures the relationships between entity nodes. Here, $K$ represents the number of entities, and $d_g$ denotes the feature dimension.
To construct the psychological graph $\mathcal{G}(M, A)$, we perform the following steps:

On the one hand, we extract the features of psychological entities for each user. These psychological entities, denoted as $E_u = \{w_1, w_2, ..., w_K\}$, are defined as psychological keywords mentioned by the user, and are automatically extracted from the user posts collections through the python tool provided by LIWC (Linguistic Inquiry and Word Count, a tool that detects words indicative of psychological states including emotion, cognition, and social orientation)\footnote{\url{https://github.com/chbrown/liwc-python}}. 
These entities include words appearing in user posts that are categorized by LIWC. 
For example, in the sentence \textit{“I am happy today”}, the word \textit{“happy”} is identified as part of the LIWC category \textit{affect} and is thus included as a psychological entity in the user’s psychological graph.
Next, we employ GloVe \cite{pennington-etal-2014-glove} to convert these entities into node features. 

Formally, the feature vector for each entity node is acquired using GloVe word embeddings, as shown in Equation (\ref{eq:GloVe}):

\begin{equation}
    \tilde{e_i} = \mathrm{GloVe}(w_i) \in \mathbb{R}^{1 \times d_g},
\label{eq:GloVe}
\end{equation}
where $w_i$ represents the $i$-th entity node of the user, $\mathrm{GloVe}(\cdot)$ denotes GloVe word vector representation.
Finally, we obtain the node feature matrix $M = \{e_1, e_2, \ldots, e_K\}$.

On the other hand, we establish edges between the extracted entity nodes based on co-categories in the LIWC knowledge base to capture the relationships between them. 
Specifically, when two entities share at least one LIWC category, an edge is created between them. 
For example, \textit{“happy”} and \textit{“sad”} both belong to the category \textit{affect}, so an edge is established between these nodes in the graph.
This process results in an adjacency matrix $A \in \mathbb{R}^{K \times K}$, where $A_{ij}=1$ signifies an edge between node $i$ and $j$, and $A_{ij}=0$ indicates no connection. This process can be formulated as shown in Equation (\ref{eq:adj construction}):

\begin{equation}
A_{ij}=\left\{\begin{aligned}
1 & , C(e_i) \cap C(e_j) \ne \emptyset, \\
0 & , \text{otherwise}.
\end{aligned}\right.
\label{eq:adj construction}
\end{equation}
where $C(\cdot)$ represents the function that retrieves the set of LIWC categories for a psychological entity.

\subsection{Dual Personality Encoding Layer}

Dual Personality Encoding Layer is the core of DEN, which enables us to learn the long-term stable user personality trait representations by analyzing the usage patterns of psychological entities based on the psychological graph. Simultaneously, it obtains the current short-term user personality state representations from dynamic context of users' individual posts. This layer allows for a comprehensive understanding of the user's personality by considering both the enduring traits and the transient states expressed in their posts.

Dual Personality Encoding Layer consists of three core sub-modules: Long-term Personality Encoding, Short-term Personality Encoding, and Bi-directional Interaction. In the following sections, we will provide detailed explanations of these three key components.

\subsubsection{Long-term Personality Encoding}

In this module, we utilize patterns of users' usage of psychological entities based on the psychological graph to model their long-term personality trait representation. Specifically, we leverage the Graph Convolutional Neural Network (GCN) \cite{kipf2016semi} to propagate the entity information within the psychological graph $\mathcal{G}(M, A)$ to capture users' usage patterns of psychological entities.

Formally, in the multiple layers of GCN, each layer utilizes the output $M^{l-1}$ from the previous layer as input to learn the current layer representation $M^l$. The learning formula for each layer of GCN is given by Equation (\ref{eq:GCN_layer1}):

\begin{equation}
\begin{aligned}
M^{l} &= \sigma\left(\hat{A} M^{l-1} W^{l}\right), \\
\hat{A} &= D^{-\frac{1}{2}} A D^{-\frac{1}{2}}, \\
\end{aligned}
\label{eq:GCN_layer1}
\end{equation}
where $\sigma(\cdot)$ represents the LeakyReLU activation function, and $W^{l} \in \mathbb{R}^{d \times d}$ is the learnable transformation matrix. $D$ is the degree matrix with $D_{ij} = \sum_j A_{ij}$, and $\hat{A}$ is the normalized adjacency matrix. 

Depending on the number of convolution layers used, GCN can aggregate information about direct neighboring nodes (using one convolution layer) or up to K-hop neighbor nodes (if K layers are stacked). For more detailed information about GCN, please refer to \cite{kipf2016semi}.

In this work, we utilize two layers of GCN, the first layer takes the initialized entity representations $M \in \mathbb{R}^{K \times d_{g}}$ as input, and the final layer outputs $M^l= \{e_1^l, e_2^l, \ldots, e_K^l\}$. Through this process, the GCN effectively captures users' usage patterns of psychological entities, resulting in enriched and informative long-term personality representations $M^l \in \mathbb{R}^{K \times d}$.

\subsubsection{Short-term Personality Encoding}

In the Short-term Personality Encoding module, the user's short-term personality states is represented by the dynamic context of users' individual posts.
To obtain the contextual representation of each post, we utilize the pre-trained language model BERT \cite{kenton2019bert} to acquire the contextual representation of each post $p_i$. The contextual representation of each post, denoted as $p_i$, is acquired as follows:

\begin{equation}
    h_i = \mathrm{BERT}(p_i) \in \mathbb{R}^{L \times d},
\label{eq:BERT_user}
\end{equation}
where $p_i$ represents the $i$-th post of the user, $\mathrm{BERT}(\cdot)$ refers to the pre-trained BERT's hidden state of the ``[CLS]'' token in each post text, and $d$ denotes the hidden dimension of BERT.

The ``[CLS]'' token in BERT is added as the first token of each post and is used to capture the overall content and context of the entire input text.

Consequently, we gather the representations of the ``[CLS]'' token, denoted as $h_{i}^l = {t} _{0i}^{l} \in \mathbb{R}^{1 \times d}$, from each post as the final short-term representations $H^l = \{h_0^l, h_1^l,..., h_N^l\} \in \mathbb{R}^{N \times d}$. 
This way, Short-term Personality Encoding effectively captures the user's informative post representations, leading to enriched and informative short-term state representations.

\subsubsection{Bidirectional Interaction}

To capture dynamic development while learning a relatively stable personality, we enhance the interaction between two personality representations, allowing them to mutually improve each other. This is achieved through the introduction of a bi-directional interaction module.

Specifically, we first establish a bipartite graph, denoted as $\widetilde{\mathcal{G}}(\widetilde{M}, \widetilde{A})$. The node matrix $\widetilde{M} \in \mathbb{R}^{(K+N) \times d}$ is the concatenation of long-term personality representations $M^l$ and short-term representations $H^l$. The adjacency matrix $\widetilde{A}$ represents the relationship between user posts and entities, with $\widetilde{A}_{ij}$ equaling 1 if entity $e_i$ appears in post $p_j$, and 0 otherwise. 

Based on the bipartite $\widetilde{\mathcal{G}}(\widetilde{M}, \widetilde{A})$, we utilize another GCN to refine the representation of the user's long-term and short-term aspects, which is formalized as follows:

\begin{equation}
\begin{aligned}
\widetilde{M}^{l} &= \sigma\left(\hat{\widetilde{A}} \widetilde{M}^{l-1} W^{l}\right), \\
\hat{\widetilde{A}}& =D^{-\frac{1}{2}} \widetilde{A} D^{-\frac{1}{2}}, \\
\end{aligned}
\label{eq:GCN_layer2}
\end{equation}

Finally, we can obtain the new long-term personality representations $\widetilde{M}^l= \{\widetilde{e}_{1}^{l}, \widetilde{e}_{2}^{l}, \ldots, \widetilde{e}_{K}^{l}\}$ and new short-term personality representations $\widetilde{H}^l= \{\widetilde{h}_{1}^{l}, \widetilde{h}_{2}^{l}, \ldots, \widetilde{h}_{N}^{l}\}$. This bi-directional interaction enhances the interaction between them and enables us to effectively capture dynamic development while learning a relatively stable personality.

\begin{figure*}[!ht]
\centering  
\subfigure[Statistical characteristics of kaggle dataset.]{   
\begin{minipage}{8.5cm}
\centering    
\includegraphics[scale=0.25]{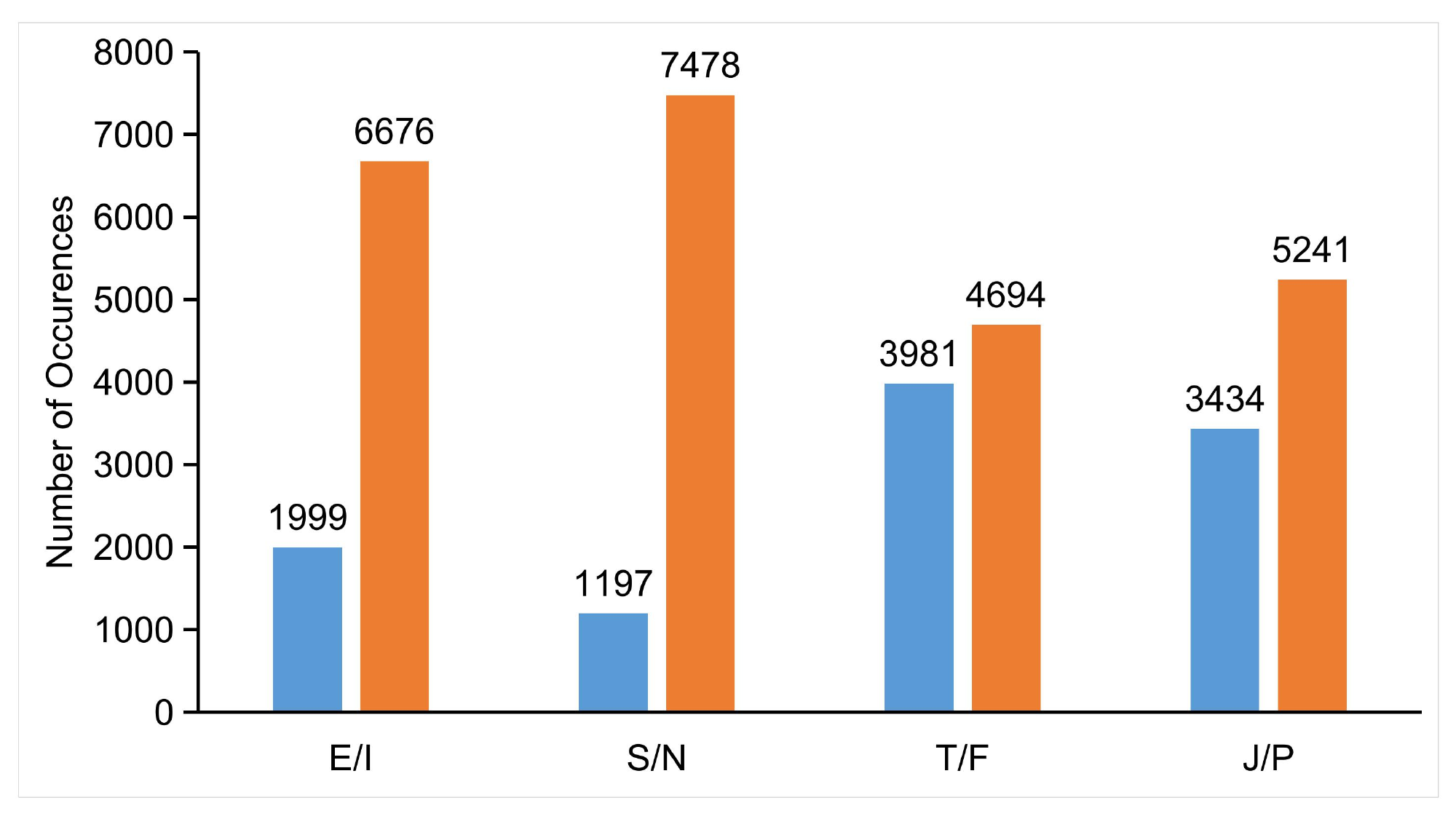}  
\end{minipage}
}
\subfigure[Statistical characteristics of pandora dataset.]{ 
\begin{minipage}{8.5cm}
\centering    
\includegraphics[scale=0.25]{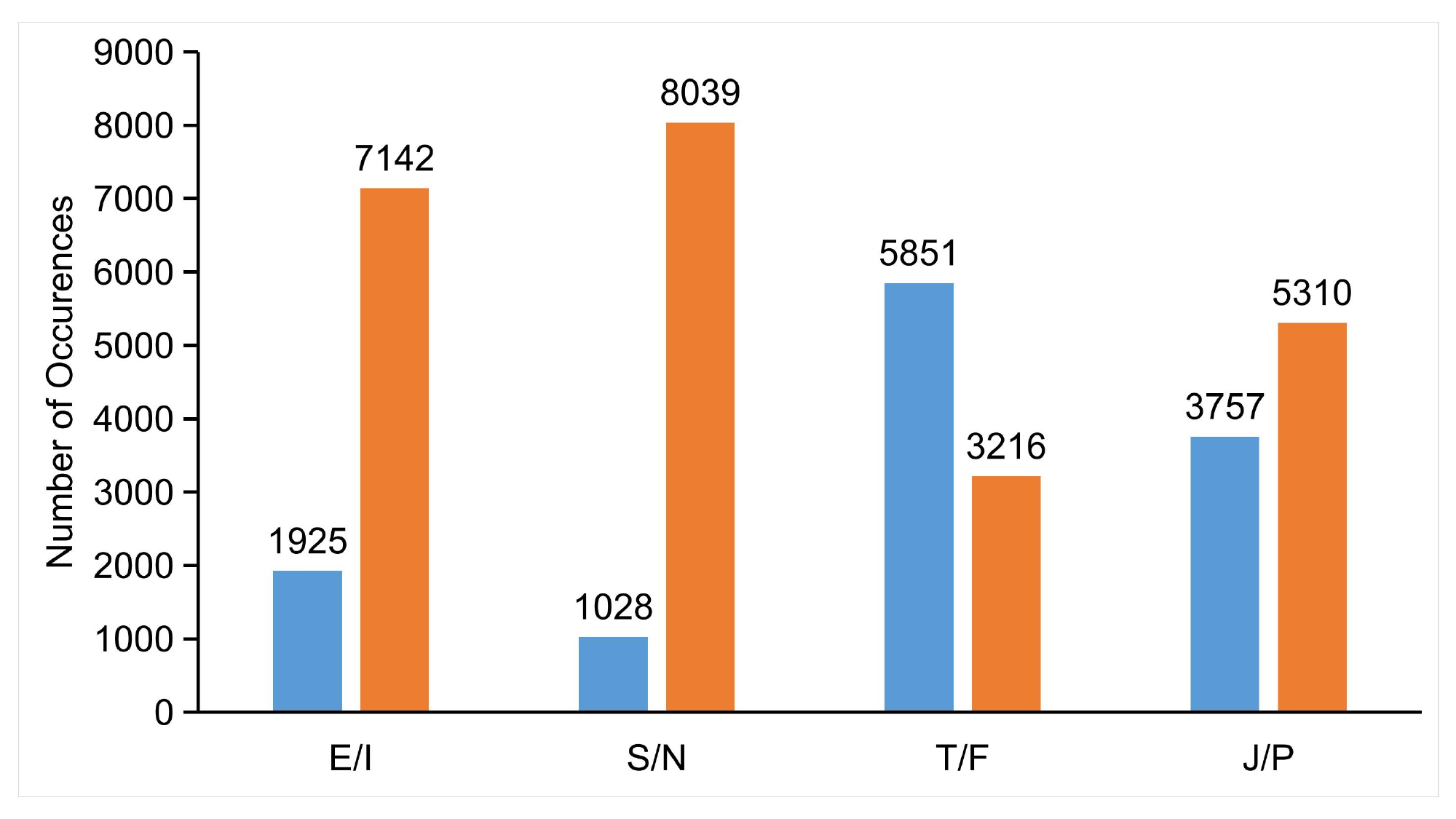}
\end{minipage}
}
\caption{Statistical charts for personality types in Kaggle and Pandora datasets.}    
\label{fig:dataset}    
\end{figure*}

\subsection{Aggregation Layer}

In Aggregation Layer, we will generate the final user representation by aggregating the long-term trait representations and short-term state representations extracted by the Dual Personality Encoding Layer. Finally, we will predict the personality characteristics based on the acquired user representation.

Since each user has multiple posts and entities, we need to aggregate the long-term and short-term representations to obtain a single user vector for personality detection. To achieve this, we first adopt a simple yet effective approach of average pooling, similar to \cite{keh2019myers, jiang2020automatic}, to refine the long-term and short-term representations into two low-dimensional sparse user vectors, $U_l \in \mathbb{R}^{1 \times d}$ and $U_s \in \mathbb{R}^{1 \times d}$:

\begin{equation}
\begin{aligned}
U_l &= \frac{1}{N} \sum_{i} \widetilde{e}_{i}^{l} \\ 
U_s &= \frac{1}{K} \sum_{i} \widetilde{h}_{i}^{l}.
\end{aligned}
\end{equation}

Next, we apply a gated fusion mechanism to control the effects of the long-term trait representation and short-term state representation on the final personality prediction. This gated fusion mechanism allows us to flexibly learn from both representations and make the final personality prediction:

\begin{equation}
\begin{aligned}
U=\alpha U_l+\left(1-\alpha\right) U_s, \\
\alpha=\sigma\left(\left[U_l ; U_s\right] W+b\right),
\end{aligned}
\label{eq:gate}
\end{equation}
where ${W} \in \mathbb{R}^{2 d \times 1}$ is a trainable weight matrix and $b \in \mathbb{R}$ is a bias term.
Based on the final user representation $U$, we employ a linear transformation followed by a softmax function to predict each personality characteristics:

\begin{equation}
\begin{aligned}
    \hat{y} = \text{Softmax}(UW_u + b_u), \\
\end{aligned}
\end{equation}
where $W_u \in \mathbb{R}^{d \times 2}$ and $b_u \in \mathbb{R}^{2}$ are both trainable parameters.

\subsection{Objective Function}

During training, the model aims to optimize its parameters by minimizing the binary cross-entropy loss, which measures the difference between the predicted probability distribution and the actual personality labels for all $T$ personality characteristics. The loss function is defined as:

\begin{equation}
    \mathcal{L}=-\sum_{t=1}^Ty_t\log\left(\hat{y}_t\right)+(1-y_t)\log\left(1-\hat{y}_t\right),
\end{equation}
where $y_t \in \{0, 1\}$ represents the ground truth personality characteristic label, and $\hat{y}_t \in \{0, 1\}$ is the predicted logits for the $t$-th personality characteristic. The binary cross-entropy loss function is particularly well-suited for binary classification tasks like personality characteristic prediction. It enables the model to consider both the long-term stable personality of users and their dynamic changes over time, thereby improving the accuracy of personality detection based on user-generated content.

\section{Experiments}

\subsection{Datasets}

In the field of personality detection, the two most popular personality models are the Big Five \cite{digman1990personality} and MBTI (Myers-Briggs Type Indicator) \cite{myers1987introduction}.

Due to our psychological theoretical analysis being based on the MBTI personality model, and the ease of obtaining MBTI labels making it more applicable for developing a personality detection model toward social media \cite{celli2018big}, we have chosen to utilize two widely used MBTI personality detection datasets for our research. 
The MBTI personality model classifies personality types into four characteristics include \textit{Extroversion} vs. \textit{Introversion} (E/I), \textit{Sensing} vs. \textit{Intuition} (S/N), \textit{Thinking} vs. \textit{Feeling} (T/F), and \textit{Judging} vs. \textit{Perception} (J/P).

We utilized the publicly available Kaggle MBTI and Pandora MBTI datasets for our evaluations.
The Kaggle dataset\footnote{\url{https://www.kaggle.com/datasets/datasnaek/mbti-type}} consists of 8675 users, each contributing an average of 47 posts. The Pandora dataset\footnote{\url{https://psy.takelab.fer.hr/datasets/all/}} contains 9067 users, each contributing an average of 96 posts. 

We shuffled the datasets and divided them into a 6:2:2 split for training, validation, and testing, respectively. Consistent with prior studies \cite{jiang2020automatic, yang2023orders}, we removed all words matching any personality label to prevent information leaks.
Detailed statistics of the Kaggle and Pandora datasets are presented in \ref{fig:dataset}. 

From Figure \ref{fig:dataset}, it is evident that both datasets exhibit varying degrees of distribution imbalance across different personality dimensions. Therefore, to evaluate the performance on each personality characteristic, we employed the Macro-F1 metric. And the average Macro-F1 of all the personality characteristics was used to assess the overall performance. 
Statistical significance of pairwise differences for DEN against the best baseline is determined by the t-test ($p < 0.05$).

\subsection{Baseline Methods}

We compare DEN with the following state-of-the-art models in the personality recognition task. 

\textbf{SVM} \cite{cui2017survey} \textbf{and XGBoost} \cite{tadesse2018personality}: Employs SVM or XGBoost to classify personality characteristics based on the TF-IDF features extracted from the document, which is created by concatenating user-generated posts.

\textbf{BiLSTM} \cite{tandera2017personality}: Encodes each post using a Bi-LSTM model with GloVe word embeddings and obtains the user representation by averaging the post representations.

\textbf{AttRCNN} \cite{xue2018deep}: Utilizes a CNN-based aggregator to obtain the user representation while incorporating psycholinguistic knowledge by concatenating Linguistic Inquiry and Word Count (LIWC) features.

\textbf{BERT} \cite{keh2019myers}: Fine-tunes BERT to encode individual posts, with the user representation derived by mean pooling over the post representations.

\textbf{SN + Attn} \cite{lynn2020hierarchical}: Utilizes a hierarchical attention network, employing a GRU with word-level attention to encode each post and another GRU with post-level attention to generate the user representation.

\textbf{Trans-MD} \cite{yang2021multi}: Sequentially encodes the posts using BERT and stores them in memory, enabling posts to access the information of former ones.

\textbf{TrigNet} \cite{yang2021psycholinguistic}: Constructs a tripartite graph with three types of nodes: post, word, and category, and aggregates post information using a graph neural network.

\textbf{D-DGCN} \cite{yang2023orders}: Employs a dynamic graph neural network to model correlations within user posts and aggregates these representations based on the graph structure.

\textbf{LLMs}: Use task-specific prompts \cite{yang2023psycot} to guide large language models (including DeepSeek V3 and GPT-4o) in inferring user personality based on their posts.

\textbf{PsyCoT} \cite{yang2023psycot}: Guides LLMs to simulate the process of completing psychological questionnaires through multi-turn dialogue for user personality prediction.

\subsection{Implementation Details}

The implementation of all models was carried out using the PyTorch framework and trained on GeForce RTX 3080 GPUs. During our experiments, we utilized pre-trained BERT models from Hugging Face\footnote{\url{https://huggingface.co}}. Specifically, we employed the BERT-BASE model, which consists of 12 network layers, a hidden layer dimension of 768 ($d=768$), and 12 attention heads.

To ensure consistency with prior research \cite{yang2023orders}, we established specific parameters for the Kaggle and Pandora datasets, limiting the maximum number of posts to 50 for Kaggle ($N=50$) and 100 for Pandora ($N=100$), with a standardized maximum post length of 70 ($L=70$) characters.

For training, we employed the Adam optimizer \cite{kingma2014adam} with a learning rate of 2e-5 for the pre-trained BERT in the Short-term Personality Encoding layer and 2e-3 for the other components of the models. Mini-batch size was set to 32. The training process continued for a maximum of 10 epochs, and the model exhibiting the best performance on the validation set was chosen for testing. To maintain result consistency and minimize experimental variations, we set all random seeds used in the experiments to 1.

\begin{table}
\small
\centering
\renewcommand{\arraystretch}{1.2}
\begin{tabular}{rccccc}
\toprule[1.5pt]
\textbf{Approach} & \textbf{E/I} & \textbf{S/N} & \textbf{T/F} & \textbf{J/P} & \textbf{Average} \\
\midrule[1pt]
SVM & 53.34 & 47.75 & 76.72 & 63.03 & 60.21 \\
XGBoost & 56.67 & 52.85 & 75.42 & 65.94 & 62.72 \\
BiLSTM & 57.82 & 57.87 & 69.97 & 57.01 & 60.67 \\ 
AttRCNN & 59.74 & 64.08 & 78.77 & 66.44 & 67.25 \\
BERT & 65.02 & 59.56 & 78.45 & 65.54 & 67.14 \\
SN + Attn & 62.34 & 57.08 & 69.26 & 63.09 & 62.94 \\
Trans-MD & 66.08 & \textbf{69.10} & 79.19 & 67.50 & 70.47 \\
TrigNet & \uline{69.54} & \uline{67.17} & 79.06 & \uline{67.69} & \uline{70.86} \\
D-DGCN & 67.37 & 64.47 & \uline{80.51} & 66.10 & 69.61 \\

DeepSeek V3 & 63.31 & 54.88 & 74.20 & 50.48 & 60.72 \\
GPT-4o & 64.59 & 60.74 & 74.92 & 50.18 & 62.61 \\
PsyCoT & 66.56 & 61.70 & 74.80 & 57.83 & 65.22 \\
\midrule[1pt]
DEN(Ours) & \textbf{69.95*} & 66.39 &  \textbf{80.65*} & \textbf{69.02*} & \textbf{71.50*} \\
\bottomrule[1.5pt]
\end{tabular}
\caption{Overall Macro-F1(\%) results of DEN and baseline models on the Kaggle dataset, where the best results are shown in bold, and the second-best are underlined. Significant improvements over the best baseline are marked with * ($p < 0.05$), as determined by a paired t-test.}
\label{tb:Kaggle Results}
\end{table}

\section{Overall Results}
\label{sec: overall_results}
The overall results for DEN and the baseline models are presented in Tables \ref{tb:Kaggle Results} and \ref{tb:Pandora Results}. These results demonstrate that our proposed DEN outperforms other models, achieving the highest average F1 score on both the Kaggle and Pandora datasets. 
Notably, DEN achieves a relative improvement of 2\% on the Pandora dataset compared to the previous state-of-the-art models, underscoring its effectiveness in personality detection.

This success can be attributed to several key factors. First, the long-term Personality Encoding effectively models users' long-term personality representations by analyzing patterns in their usage of psychological entities. Second, the short-term Personality Encoding efficiently captures users' short-term representations by modeling the contextual information of each user's posts. Most importantly, the bi-directional interaction module enhances the interaction between long-term and short-term personality representations, enabling them to mutually improve each other. Through these processes, DEN can simultaneously consider both the long-term stable personality of users and their dynamic changes over time, which ultimately enhances its performance in personality detection. 

\begin{table}
\small
\centering
\renewcommand{\arraystretch}{1.2}
\begin{tabular}{rccccc}
\toprule[1.5pt]
\textbf{Approach} & \textbf{E/I} & \textbf{S/N} & \textbf{T/F} & \textbf{J/P} & \textbf{Average} \\
\midrule[1pt]
SVM & 44.74 & 46.92 & 64.62 & 56.32 & 53.15 \\
XGBoost & 45.99 & 48.93 & 63.51 & 55.55 & 53.50 \\
BiLSTM & 48.01 & 52.01 & 63.48 & 56.12 & 54.91 \\ 
AttRCNN & 48.55 & 56.19 & 64.39 & 57.26 & 56.60 \\
BERT & 56.60 & 48.71 & 64.70 & 56.07 & 56.52 \\
SN + Attn & 54.60 & 49.19 & 61.82 & 53.64 & 54.81 \\
Trans-MD & 55.26 & \textbf{58.77} & \uline{69.26} & \textbf{60.90} & \uline{61.05} \\
TrigNet & 56.69 & 55.57 & 66.38 & 57.27 & 58.98 \\
D-DGCN & \uline{58.28} & 55.88 & 68.50 & 57.72 & 60.10 \\
\midrule[1pt]
DEN(Ours) & \textbf{60.86*} & \uline{57.74} & \textbf{71.64*} & \uline{59.17} & \textbf{62.35*} \\
\bottomrule[1.5pt]
\end{tabular}
\caption{Overall Macro-F1(\%) results of DEN and baseline models on the Pandora dataset.}
\label{tb:Pandora Results}
\end{table}

It's worth noting that while Trans-MD \cite{yang2021multi} and TrigNet \cite{yang2021psycholinguistic} were not explicitly designed to consider personality development as a psychological characteristic, their designs inadvertently incorporated this aspect into their frameworks, and as a result, they performed exceptionally well in personality detection. Trans-MD is designed to access all users' posts simultaneously when encoding the current document. This design unintentionally involves the long-term information into the short-term encoding process. The TrigNet model, with its tripartite graph structure, takes into account user posts and dynamically updates user representations using psychological entities and categories as intermediaries. Indeed, their inadvertent incorporation of this aspect into their designs turned out to be one of the key factors contributing to their outstanding performance in personality detection. This further supports our argument that simultaneously considering both users' long-term stable personality and their dynamic changes over time can improve personality detection results.

However, we also observed that all models, including DEN, perform relatively poorly on the \textbf{Sensing/Intuition (S/N)} personality dimension. Moreover, DEN’s advantage in this dimension is less pronounced compared to other dimensions. We hypothesize that this is due to the highly imbalanced nature of personality detection datasets, with the \textbf{S/N} dimension being particularly skewed. As DEN integrates both long-term and short-term modeling, it may have inadvertently amplified these biases. To address this issue, future work could explore strategies such as data augmentation, class rebalancing, or the use of focal loss to mitigate the impact of data imbalance.

\section{Detailed Analysis}
The overall results presented above have demonstrated the effectiveness of our DEN model as a whole. To gain deeper insights and understand the impact of each key module in DEN, we conducted a series of experiment on various variants of DEN, and the general results are shown in Table \ref{tb:Ablation Results}. It can be observed that removing any part of the model leads to a performance decrease in DEN. The detailed findings and analysis will be discussed in the following sections.

\begin{table}[ht]
\small
\renewcommand{\arraystretch}{1.2} 
\centering 
\begin{tabular}{lccccc}
\toprule[1.5pt]
\textbf{Approach} & \textbf{E/I} & \textbf{S/N} & \textbf{T/F} & \textbf{J/P} & \textbf{Average}\\
\midrule[1pt]
DEN\textsubscript{-short} & 43.56 &	46.15 &	67.56 &	57.89 &	53.79\\
DEN\textsubscript{-long} & 65.02 & 59.56 & 78.45 &	65.54	& 67.14\\
DEN\textsubscript{-gcn} & 67.42 & 63.76 &	78.70 &	\uline{66.52} &	69.10 \\
DEN\textsubscript{-inter} & \uline{68.06} &	64.39 &	78.66 &	66.12 &	69.31 \\
DEN\textsubscript{-liwc} & 67.68 &	\textbf{66.48} &	\uline{78.85} &	66.47 &	\uline{69.87}\\
\midrule[1pt]
DEN(Ours) & \textbf{69.95} & \uline{66.39} &  \textbf{80.65}   & \textbf{69.02} & \textbf{71.50}
\\
\bottomrule[1.5pt]
\end{tabular}
\caption{Results of variants of DEN model in Macro-F1(\%) on the test set of Kaggle dataset.}
\label{tb:Ablation Results}
\end{table}

\subsection{Effect of Long-term Personality Encoding}

\begin{figure}[!ht]
\centering
    \includegraphics[scale=0.25]{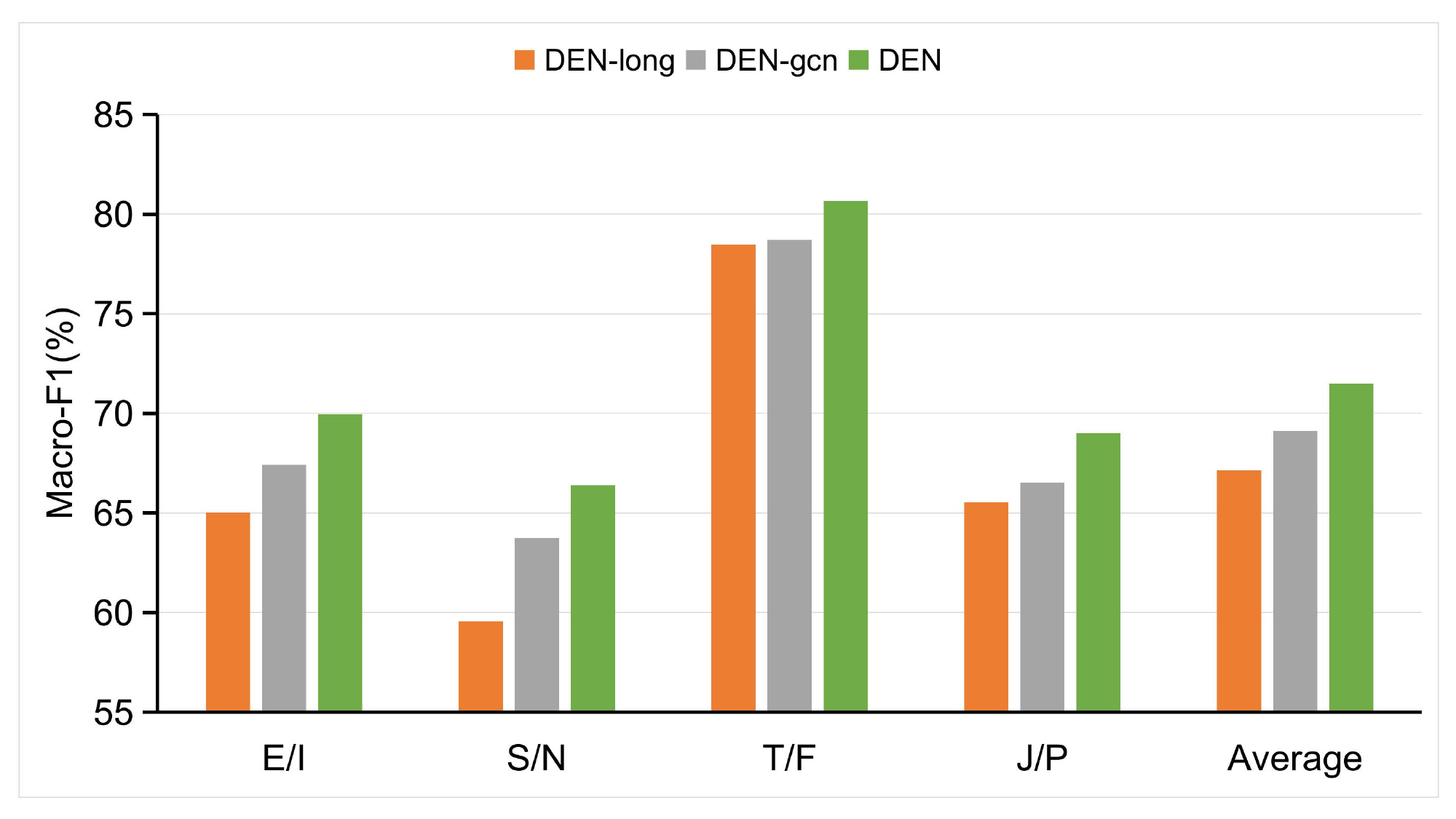}
\caption{Effect of Long-term Personality Encoding}
\label{fig:ablation1}
\end{figure}

In the variant DEN\textsubscript{-long}, we removed the entire Long-term Personality Encoding layer and no long-term information was incorporated into the modeling process, resulting in a 4.36\% decrease in performance. This indicates the crucial importance of considering long-term information in personality detection. 
In the variant DEN\textsubscript{-gcn}, we removed the graph convolutional neural network module in the Long-term Personality Encoding layer but still fed the raw psychological entity representations into the Bi-directional Interaction module, which led to a 2.4\% decrease in performance. This emphasizes the significance of modeling long-term user information for personality detection. 
These provide evidence that our DEN effectively utilizes graph convolutional neural networks to analyze patterns in users' usage of psychological entities, allowing it to efficiently acquire users' long-term personality representations and improve personality detection performance.

\subsection{Effect of Short-term Personality Encoding}

\begin{figure}[!ht]
\centering
    \includegraphics[scale=0.25]{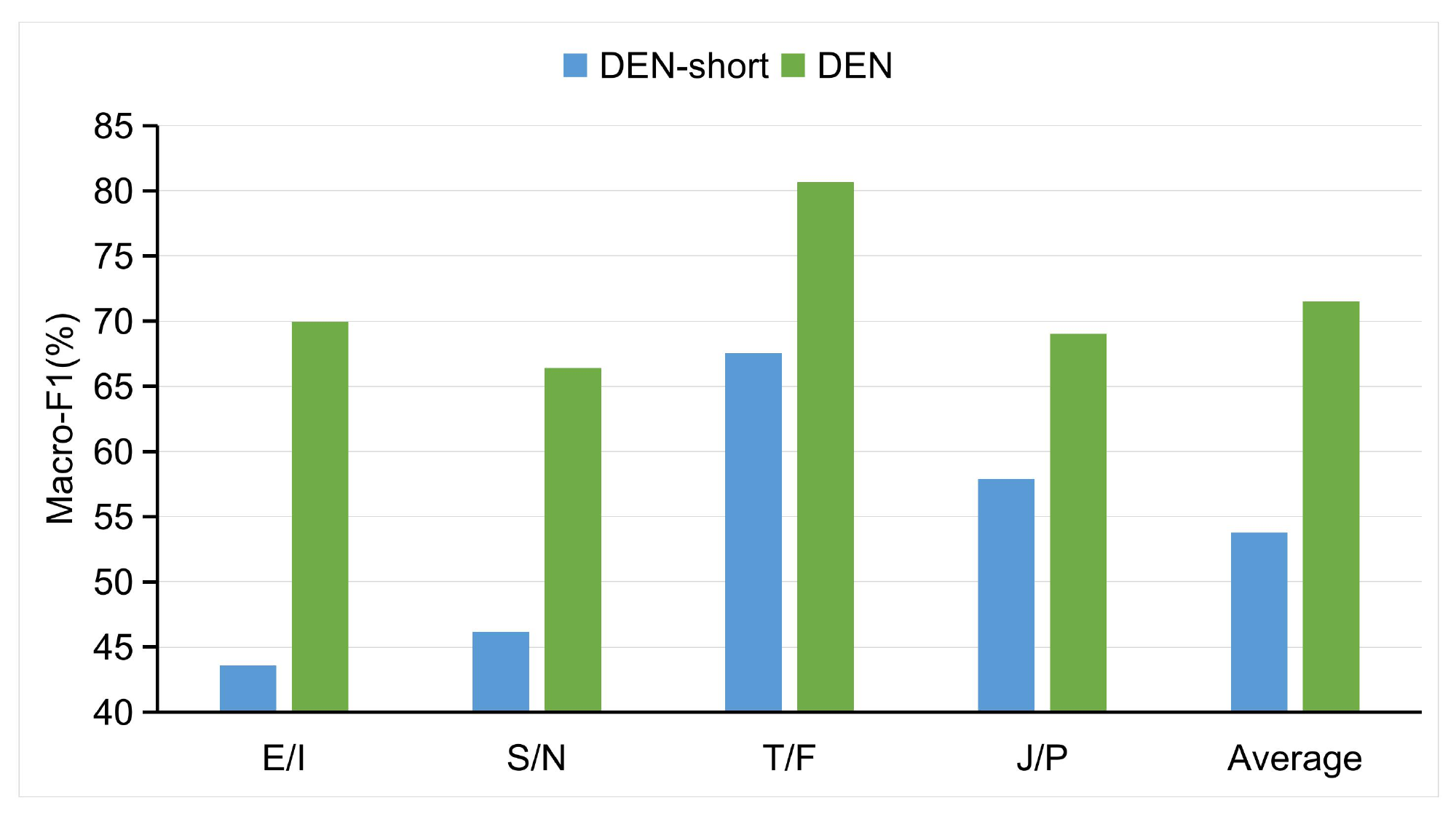}
\caption{Effect of Short-term Personality Encoding}
\label{fig:ablation2}
\end{figure}

When the short-term Personality Encoding module is removed, denoted as DEN\textsubscript{-short}, the performance drops dramatically, almost to an unusable level. This is because, without the short-term Personality Encoding module, the model only considers users' habits in using psychological entities, leading to a significant loss of semantic information. While these habits provide some level of user modeling, the absence of post-level semantic summary vectors makes it difficult to predict user personalities. Furthermore, the model is hindered by a scarcity of parameters at this point, which poses challenges in accurately fitting the data.
Therefore, similar to previous work, DEN utilizes pre-trained language models and the user's textual context information to model the user's short-term representations, enriching user information and thus enhancing personality detection performance.

\subsection{Effect of Bi-directional Interaction}

When the Bi-directional Interaction module is removed in DEN\textsubscript{-interaction}, the long-term trait representations and short-term state representations are directly aggregated for personality prediction. This also results in a noticeable decrease in performance, but it still outperforms DEN\textsubscript{-long} and DEN\textsubscript{-short}.

\begin{figure}[!ht]
\centering
    \includegraphics[scale=0.25]{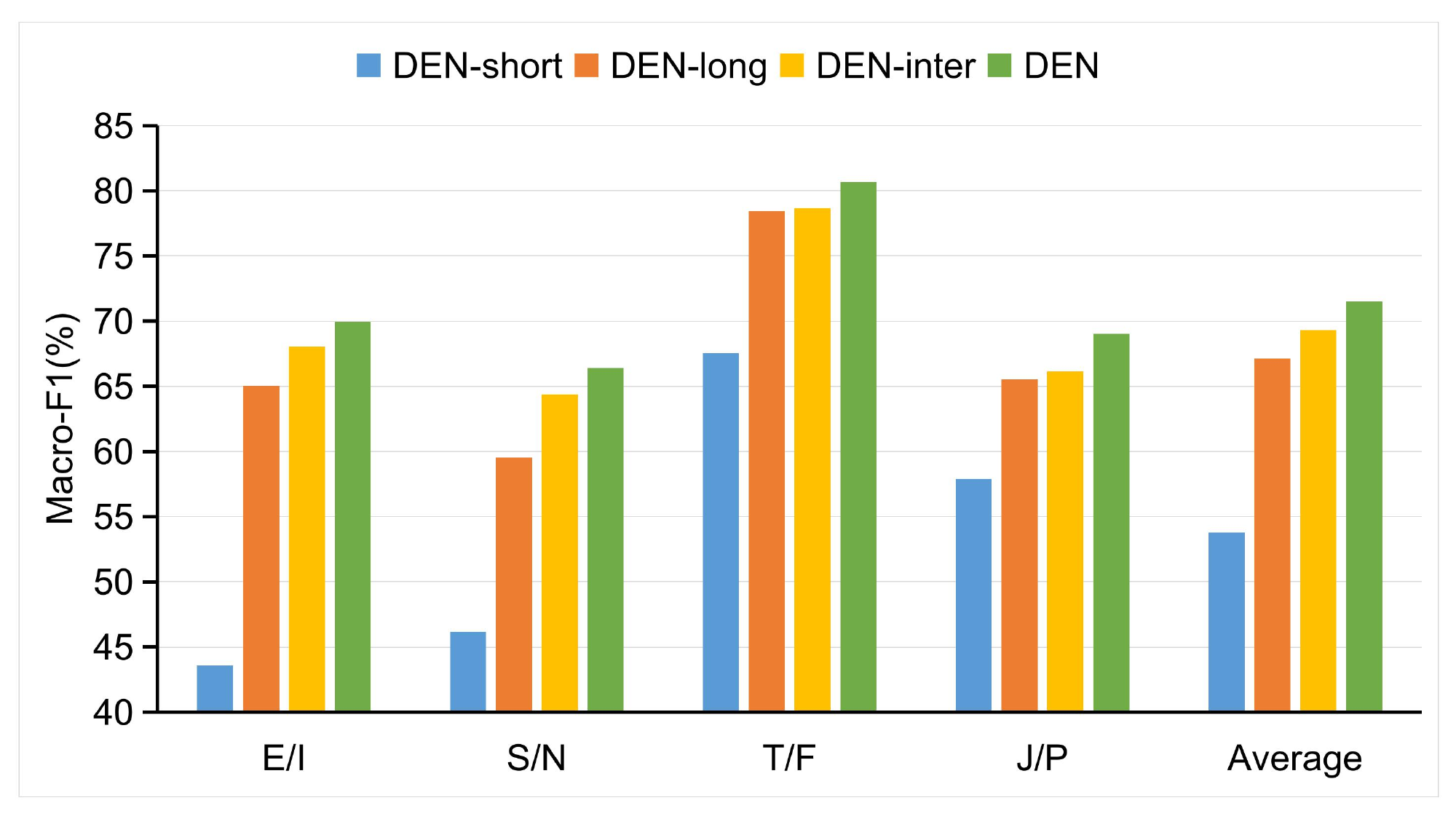}
\caption{Effect of Bi-directional Interaction}
\label{fig:ablation3}
\end{figure}

This is because both long-term and short-term personality representations already contain rich user information, but they cannot dynamically interact with each other to enhance their individual qualities. While this interaction can be indirectly achieved through subsequent updates of neural network parameters, DEN explicitly implements this functionality, leading to a performance improvement to some extent.

\subsection{Effect of External Knowledge Base}

To explore the effect of the External Knowledge Base on DEN, we configured the input of the GCN in the long-term Personality Encoding to be a fully connected graph, meaning that every pair of entities has an edge connecting them. This variant is denoted as DEN\textsubscript{-liwc}. Without the guidance of external psychological knowledge, the DEN\textsubscript{-liwc} model exhibits a noticeable decrease in performance. This indicates that DEN can effectively utilize external psychological knowledge to enhance the effectiveness of personality detection.

\begin{figure}[!ht]
\centering
    \includegraphics[scale=0.25]{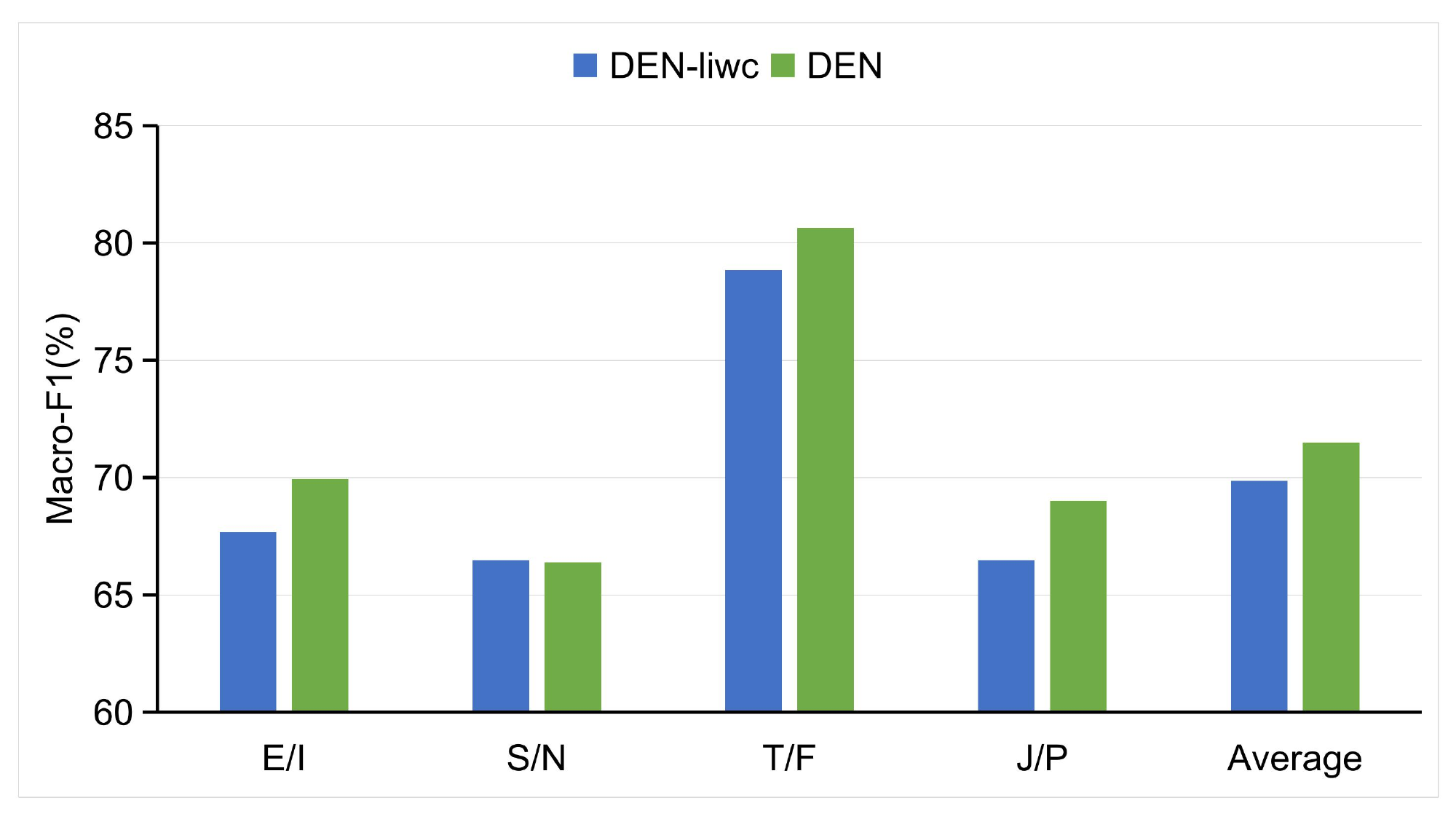}
\caption{Effect of External Knowledge Base}
\label{fig:ablation4}
\end{figure}

\begin{table*}
\renewcommand{\arraystretch}{1.2} 
\centering 
\setlength{\tabcolsep}{1.5mm}{
\begin{tabular}{ccccccccccccccccc}
\toprule[1.5pt]
\textbf{Approach} & \textbf{INFP} & \textbf{INFJ} & \textbf{INTP} & \textbf{INTJ} & \textbf{ISFP} & \textbf{ISFJ} & \textbf{ISTP} & \textbf{ISTJ} & \textbf{ENFP} & \textbf{ENFJ} & \textbf{ENTP} & 
\textbf{ENTJ} & \textbf{ESFP} & \textbf{ESFJ} & \textbf{ESTP} & \textbf{ESTJ}\\
\midrule[1pt]
Support (\%) & 21.27 & 17.18 &  15.68 & 11.12 & 3.52 & 2.19 & 4.15 & 2.07 & 8.01 & 1.84 & 8.18 & 2.42 & 0.52 & 0.52 & 1.04 & 0.29 \\

\midrule[1pt]\

D-DGCN & 53.69 & 41.09 & 46.79 & 34.43 & \textbf{25.58} & 15.39 & \textbf{33.33} & 15.39 & 32.23 & 9.88 & \textbf{39.73} & 25.32 & 0.00 & 0.00 & 6.67 & 0.00 \\
DEN(Ours) & \textbf{54.18} & \textbf{45.89} & \textbf{52.03} & \textbf{37.46} & 19.97 & \textbf{16.14} & 29.79 & \textbf{22.95} & \textbf{34.10} & \textbf{11.59} & 38.69 & \textbf{26.54} & \textbf{0.00} & \textbf{0.00} & \textbf{12.35} & \textbf{0.00} \\
\bottomrule[1.5pt]
\end{tabular}
}
\caption{Results of DEN and D-DGCN in Macro-F1(\%) on the test set of Kaggle dataset on various personality combinations. Support refers to the ratio of support sets for each personality combination.}
\label{tb:dimensional score}
\end{table*}

\subsection{Effect of Gated Fusion Mechanism}

\begin{figure}[!ht]
\centering
    \includegraphics[scale=0.25]{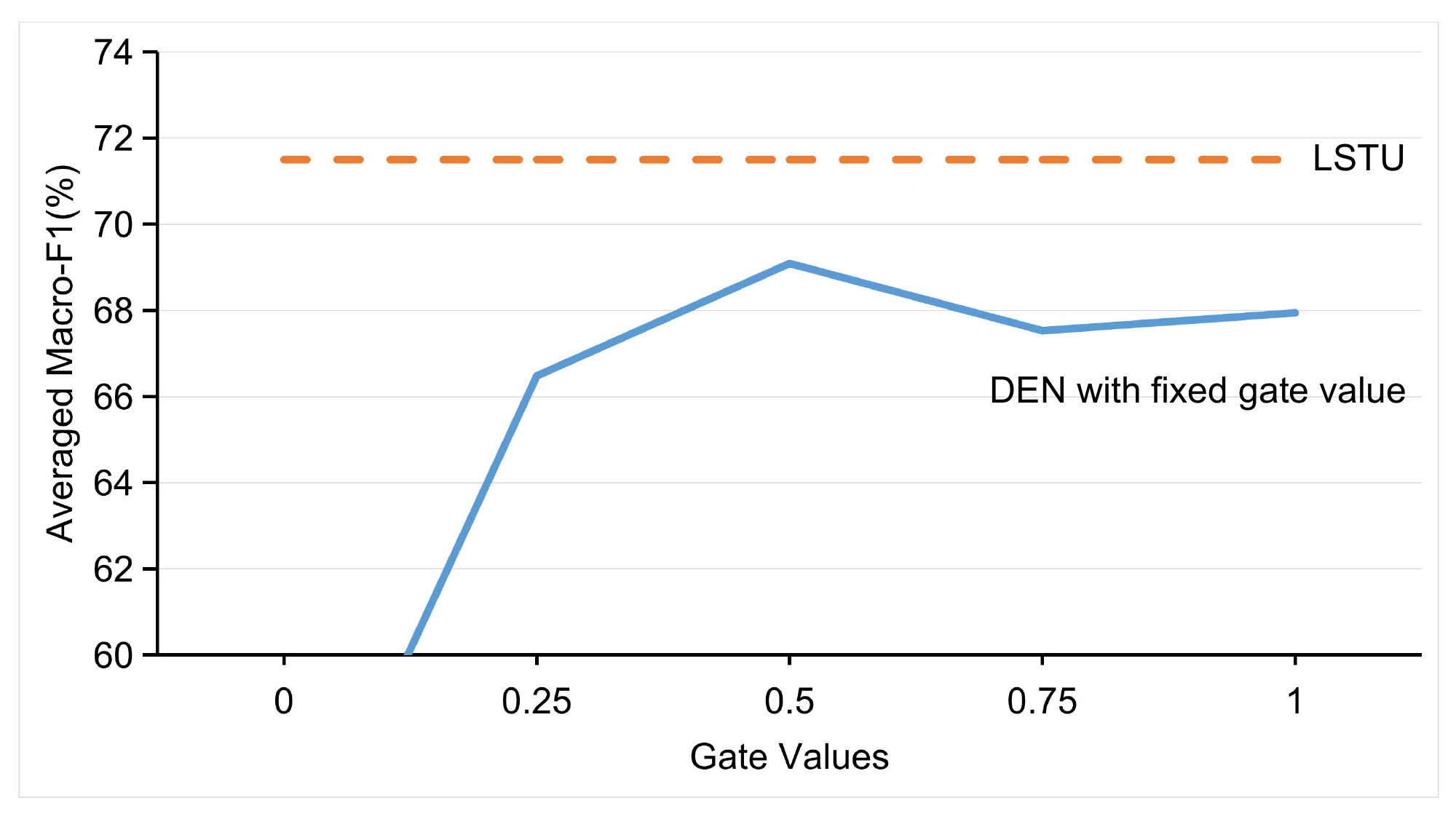}
\caption{Effect of Gated Fusion Mechanism. The dashed line represents the performance of the original LSTU model, while the solid line represents the performance of DEN's variants with different gate values.}
\label{fig:ablation5}
\end{figure}

To investigate the gated fusion mechanism in Equation (\ref{eq:gate}), we conducted experiments where we replaced the learnable gate value $\alpha$ with different fixed values. Since $\alpha$ is a continuous variable ranging from zero to one, we selected values from the set $\{0, 0.25, 0.5, 0.75, 1.0\}$. The experimental results of DEN's variants with different gate values are shown in Figure \ref{fig:ablation5}.

From the results, we observed that setting gate value to zero degrades DEN into DEN\textsubscript{-long}, resulting in poor performance, and increasing gate value generally improves the performance of DEN, but performance starts to decline once $\alpha$ exceeds a certain threshold. 
However, regardless of the chosen fixed value, the performance was consistently inferior compared to using the learnable gate.
This can be attributed to the fact that DEN's gated fusion mechanism can adaptively generate suitable gate values based on the characteristics of each user. This allows it to determine the impact of long-term and short-term personality presentations on the final personality detection specifically for each user. On the other hand, using fixed weights fails to achieve this adaptability. As a result, fixed weights that may work well for certain users may perform poorly for others.

\subsection{Effect of Number of Layers of Bi-Directional Interaction}

\begin{figure}[!ht]
\centering
    \includegraphics[scale=0.25]{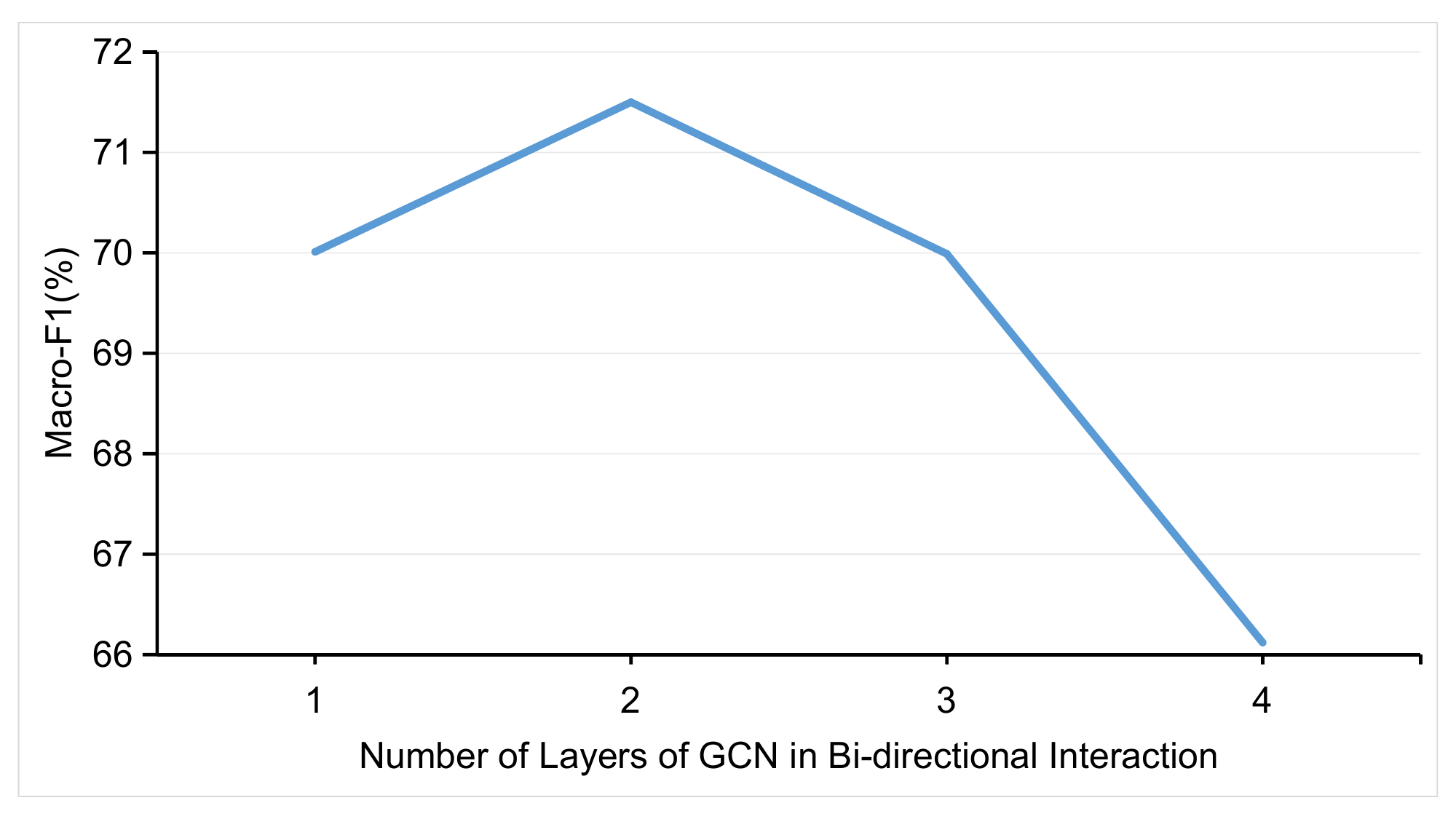}
\caption{Effect of Number of Layers of Bi-Directional Interaction}
\label{fig:ablation6}
\end{figure}

The short-term and long-term features in DEN exist in distinct spaces, and increasing the number of interaction layers may help integrate these features more effectively by reducing the gap between these spaces. To investigate the effect of the number of layers in the Bi-Directional Interaction component, we conducted a series of experiments, with the results presented in Figure \ref{fig:ablation6}.
The results show that the model achieves its best performance with two interaction layers. Beyond this point, performance declines as additional layers are added. This improvement with two layers can be attributed to the model’s ability to capture higher-order neighborhood information, enhancing the integration of long-term and short-term features. However, adding too many layers results in over-smoothing, where node representations become indistinguishable, ultimately degrading the model’s performance.
Considering these trade-offs, we adopt a two-layer network as it balances effective feature integration with avoiding over-smoothing, ensuring optimal model performance.

\subsection{Performance on Personality Combinations}

The MBTI personality framework consists of 16 unique personality combinations. 
To analyze the performance of DEN in greater detail, we report the results of DEN and D-DGCN across these combinations, as shown in Table \ref{tb:dimensional score}. 
Our findings indicate that DEN outperforms D-DGCN in 13 out of 16 combinations, demonstrating the superiority of DEN in personality detection. 
This highlights the effectiveness of modeling personality through both long-term and short-term aspects.

However, the results also reveal a significant limitation: both DEN and D-DGCN perform poorly in personality combinations with limited data. 
In combinations representing less than 1\% of the dataset, the F1-score for both models drops to 0, indicating a complete inability to make accurate predictions. 
These findings underscore the need for further research to address data scarcity in underrepresented personality combinations.

\section{Conclusion and Future Work}

In this paper, we have introduced DEN, a model that effectively integrates both the long-term stable personality traits and short-term dynamic states of users to enhance the accuracy of personality detection, considering the psychological theory that personality is relatively stable and dynamically developing.
DEN acquires long-term personality information by analyzing patterns in users' usage of psychological entities, while utilizing user post context to model short-term aspects. Furthermore, the incorporation of a novel bi-directional interaction module dynamically enhances both the long-term and short-term representations. These informative representations are then integrated for final personality detection. Our experiments provide empirical evidence for the efficacy of DEN in personality detection and highlight the advantages of considering the psychological characteristics of personality as relatively stable and dynamically developing.

In our current research, both our psychological theoretical analysis and the DEN personality detection model were based on the MBTI personality model. However, we recognize the importance of exploring the effectiveness of the DEN model on other personality models, such as the Big Five, in future work. By examining the applicability of the DEN model to different personality frameworks, we can broaden our understanding of its performance and its ability to generalize across various personality models. 
Another important direction for future research is addressing the challenge of imbalanced data in personality detection tasks. Many personality traits and combinations are underrepresented in existing datasets, which can lead to biased model performance and poor generalization, particularly for less frequent classes. While our study employed Macro-F1 as an evaluation metric to mitigate the influence of imbalance, this does not fully resolve the issue. Future work could explore advanced methods, such as data augmentation, re-sampling strategies, or loss functions like focal loss, to better address class imbalance and improve the robustness of personality detection models across all personality categories. These enhancements would contribute to more equitable and accurate personality prediction, even for underrepresented traits and combinations.

\section{Acknowledgements}
This research is supported by the National Natural Science Foundation of China (No. 62376051).

\bibliographystyle{IEEEtran}
\bibliography{DEN-TASLP}

\end{document}